\documentclass[10pt,twocolumn]{article}

\usepackage[margin=0.75in]{geometry}
\usepackage{times}
\usepackage{natbib}
\usepackage{graphicx}
\usepackage{amsmath,amssymb}
\usepackage{booktabs,multirow}
\usepackage[table]{xcolor}
\usepackage{subcaption}
\usepackage{url}
\usepackage{hyperref}
\hypersetup{colorlinks=true,allcolors=blue,breaklinks=true}

\title{Confidently Wrong: How False Certainty Shields Against Model Inversion, and a Cheaper Alternative}
\author{%
\begin{tabular}{ccc}
\begin{minipage}[t]{0.29\textwidth}\centering
\textbf{Yuxin Zhang}\\National University of Defense Technology, Changsha, China\\\texttt{zhangyuxin@nudt.edu.cn}
\end{minipage} &
\begin{minipage}[t]{0.29\textwidth}\centering
\textbf{Xiong Peng}\\Hong Kong Baptist University\\Hong Kong, China\\\texttt{csxpeng@comp.hkbu.edu.hk}
\end{minipage} &
\begin{minipage}[t]{0.29\textwidth}\centering
\textbf{Wenjie Zhu}\\National University of Defense Technology, Changsha, China\\\texttt{zhuwenjie@nudt.edu.cn}
\end{minipage} \\
\begin{minipage}[t]{0.29\textwidth}\centering
\textbf{Long Lan}\\National University of Defense Technology, Changsha, China\\\texttt{fourth.author@example.com}
\end{minipage} &
&
\begin{minipage}[t]{0.29\textwidth}\centering
\textbf{Bo Han}\\Hong Kong Baptist University\\Hong Kong, China\\\texttt{bhanml@comp.hkbu.edu.hk}
\end{minipage}
\\
\end{tabular}}
\date{}

\definecolor{tblHeader}{gray}{0.91}
\definecolor{tblSection}{gray}{0.95}
\definecolor{tblHighlight}{gray}{0.91}
\definecolor{headergray}{gray}{0.92}
\definecolor{groupgray}{gray}{0.96}
\definecolor{highlightgray}{gray}{0.93}
\newcommand{\bad}[1]{{\scriptsize\textcolor{red}{(#1)}}}
\newcommand{\good}[1]{{\scriptsize\textcolor{green!60!black}{(#1)}}}
\newcommand{\err}[1]{\ensuremath{{\scriptstyle\pm #1}}}

\begin{document}
\maketitle
\begin{abstract}
Model inversion attacks (MIAs) reconstruct private training data from trained models, motivating a range of training-based defenses. These defenses are typically judged by how much they reduce attack accuracy, leaving open why they work and what they cost. We study these questions through the lens of out-of-distribution (OOD) confidence. We find that three representative defenses built on different principles, namely BiDO, MIDRE, and Neg.LS, all exhibit false certainty: they assign high confidence to natural OOD samples, such as faces of unseen individuals. Within each defense, higher OOD confidence is strongly associated with lower attack accuracy. By analyzing confidence landscapes and inversion trajectories, we show that false certainty extends high target-class confidence to regions far from the target, so the attack saturates prematurely before reaching it, and temperature-scaling interventions support this link in both directions. However, this resistance comes at a cost: the defended models confidently match unseen individuals to known identities, a reliability risk for open-set face recognition. We further identify an alternative route to inversion resistance through Input Mixup, which achieves the lowest attack accuracy in 7 of 8 settings while keeping OOD confidence at or below that of an undefended model. This demonstrates that strong inversion resistance can be achieved without incurring the OOD reliability cost associated with overconfidence-based defenses.
\end{abstract}

\section{Introduction}
\label{sec:intro}


Deep neural networks (DNNs)~\citep{dnn2,dnn1} are widely deployed in privacy-sensitive applications such as facial recognition~\citep{facenet,deepface}, medical diagnosis~\citep{skindiagnosis}, and personalized recommendation~\citep{recommend,youtube}. Releasing trained models instead of raw data may appear to protect the underlying data, yet model inversion attacks (MIAs)~\citep{mirror,kedmi,fredrikson2014,fredrikson2015,vmi,gmi} can reconstruct recognizable private training samples from model access alone, particularly in the white-box setting, where adversaries can exploit model parameters and gradients~\citep{lomma,ppdg,manifold,PPA}. To mitigate this threat, a variety of training-based defenses have been proposed~\citep{TL,Trap-MID,bido,ls,AT-AT,RE,mid}. These defenses are typically judged by a single criterion: how much they reduce attack accuracy while preserving test accuracy. This leaves two questions open. First, \emph{why} do they work? Lower attack accuracy may reflect less private information retained by the model, or an attack that is misled before it reaches the target; recent work suggests the former is not always the case~\citep{AT-AT}. Second, \emph{at what cost}? Defensive training may change model behavior in ways that test accuracy does not reveal, particularly on inputs outside the training distribution.

\begin{figure*}
    \centering
    \includegraphics[width=0.97\linewidth]{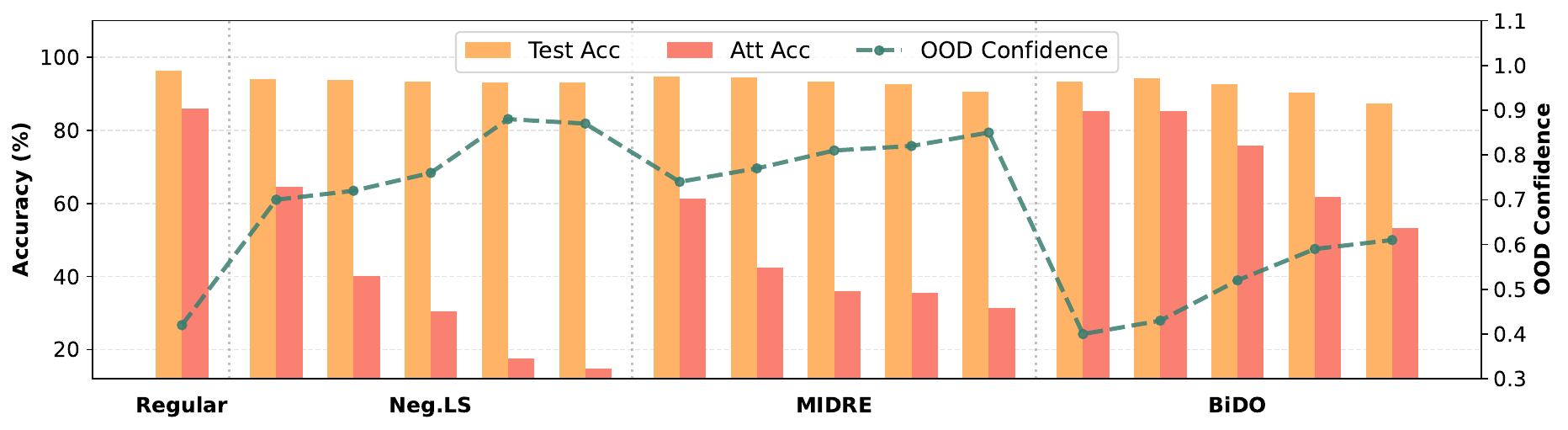}
    \caption{\textbf{OOD confidence and inversion resistance across defenses.} Each group contains models trained with different defense hyperparameters at comparable test accuracy. Within each defense, higher OOD confidence is associated with lower inversion attack accuracy.}
    \label{fig:cross-defense acc&ood}
\end{figure*}

Focusing on high-resolution MIAs, we examine three representative defenses, including BiDO~\citep{bido}, MIDRE~\citep{RE}, and Neg.LS~\citep{ls}, and find that all three exhibit markedly higher \emph{Out-of-Distribution} (OOD) confidence than an undefended model. Such OOD overconfidence has been reported for BiDO~\citep{bido,bido+}, and BiDO+~\citep{bido+} mitigates it using auxiliary OOD data, treating it as a calibration issue. We instead ask how this behavior relates to the defense effect itself. We show that it is not specific to BiDO but is shared by defenses built on different principles. Moreover, within each defense, OOD confidence is strongly and negatively correlated with attack accuracy across defense strengths at comparable test accuracy (Pearson $r\leq -0.90$; Fig.~\ref{fig:cross-defense acc&ood}, Tab.~\ref{tab:ood_correlation}).


In our setting, OOD samples are natural images whose identities lie outside the model's training classes, \emph{e.g.}, faces of individuals unseen during training, as opposed to in-distribution (ID) samples drawn from the private training data. This distinction is central to generative MIAs: the attack starts from random latent codes far from the target regions~\citep{ppdg}, \emph{i.e.}, OOD samples, and moves them toward the target identity by maximizing target-class confidence. How a model scores OOD samples therefore directly shapes the attack's path. We visualize the target-class confidence landscape over the generator's latent space and trace inversion trajectories on it (Fig.~\ref{fig:a.landscape}), revealing a sharp contrast. For the regular model, high target-class confidence is confined to a narrow region around the target identity; samples far from it receive low confidence, so the attack keeps receiving an informative signal and steadily approaches the target. For BiDO, MIDRE, and Neg.LS, high target-class confidence instead extends over broad regions far from the target. The attack enters these regions early and reaches near-maximal confidence while the reconstructions remain far from the target; as the gradient norm diminishes, further progress toward the target becomes limited (Fig.~\ref{fig:b.dynamics}; Fig.~\ref{fig:gradient_norm_and_msp}(a)). We term this \emph{premature saturation} and identify it as one mechanism behind the resistance of these three defenses. Temperature scaling~\citep{ts} analysis in Fig.~\ref{fig:TS} supports this view: softening an overconfident defended model's outputs lowers its OOD confidence, accompanied by an increase in attack accuracy from 24\% to 84\%.

However, this defense effect comes at a cost. The elevated confidence on unfamiliar inputs that contributes to misleading the attack also makes these defenses unreliable when they encounter unseen identities. In open-set face recognition, for example, an unenrolled individual may receive a near-certain prediction for a known identity (Fig.~\ref{fig:real vs ood}), potentially undermining the reliable rejection of unknown individuals in practice~\citep{li2025position,bido+,vareto2024open}. Such defenses protect training data by making the model confidently wrong on unfamiliar inputs, trading one risk for another.



    



This raises a natural question: \emph{is OOD overconfidence a necessary price for inversion resistance?} We find that it is not. Input Mixup~\citep{mixup}, a standard training technique that interpolates pairs of training inputs and their labels, provides strong inversion resistance while keeping OOD confidence at or below that of a regular model. It achieves the lowest attack accuracy in 7 of the 8 evaluated settings (Tab.~\ref{tab:overall_comparison}). On FaceScrub with ResNet-152, it reduces PPA attack accuracy from 85.93\% to 10.93\% at a cost of 1.75\% test accuracy, while lowering OOD confidence from 0.42 to 0.27 (Tab.~\ref{tab:main}). Unlike BiDO+~\citep{bido+}, it requires no auxiliary OOD data, additional loss terms, or architectural changes.

Our contributions are summarized as follows:

\begin{itemize}
    \item \textbf{OOD overconfidence across MIA defenses.} We show that BiDO, MIDRE, and Neg.LS all exhibit markedly higher OOD confidence than a regular model. Within each, it is strongly negatively correlated with attack accuracy across defense strengths (Pearson $r\leq -0.90$).

    \item \textbf{Premature saturation explains part of their inversion resistance.} By analyzing confidence landscapes and inversion trajectories, we show that these defenses extend high target-class confidence to regions far from the target identity, so attacks saturate before reaching it. Although these defenses resist inversion, this comes at the cost of confident predictions on unseen identities, a reliability risk for open-set face recognition.

    \item \textbf{Strong inversion resistance without elevated OOD confidence.} We show that Input Mixup, without auxiliary data, additional losses, or architectural changes, achieves the lowest attack accuracy in 7 of the 8 evaluated settings across 2 datasets $\times$ 2 architectures $\times$ 2 attacks. It also keeps OOD confidence at or below that of a regular model, with a test-accuracy drop of at most 4.1\%.

\end{itemize}

\section{Preliminaries}
\label{sec:pre}


\subsection{Model Inversion Attacks}

Let $T_\theta:\mathcal{X}\rightarrow\Delta^C$ be a target classifier trained on a private dataset $\mathcal{D}_{\mathrm{priv}}=\{(x_i,c_i)\}_{i=1}^N$, with $c_i\in\{1,\ldots,C\}$. Given access to $T_\theta$, a MIA aims to reconstruct private samples of a target class $c$~\citep{fredrikson2014,fredrikson2015}. Since directly optimizing inputs to maximize $P(c\mid x;T_\theta)$ yields unrealistic images, modern MIAs follow the generative framework of GMI~\citep{gmi}: a generator $G:\mathcal{Z}\rightarrow\mathcal{X}$ trained on a public dataset $\mathcal{D}_{\mathrm{pub}}$ constrains the search, and the attacker optimizes a latent code
\begin{equation}
\begin{aligned}
z^* &= \operatorname*{arg\,min}_{z}\mathcal{L}_{\mathrm{inv}}(z),\\
\mathcal{L}_{\mathrm{inv}}(z) &= \mathcal{L}_{\mathrm{id}}(T_\theta(G(z)),c) + \gamma\mathcal{L}_{\mathrm{prior}}(z;G),
\end{aligned}
\label{eq:mia_gan}
\end{equation}
where $\mathcal{L}_{\mathrm{id}}$ rewards high target-class confidence (e.g., cross-entropy or Poincar\'e loss), and $\mathcal{L}_{\mathrm{prior}}$ keeps $G(z)$ realistic.

We focus on the high-resolution attacks, PPA~\citep{PPA} and PPDG~\citep{ppdg}, which use a pretrained StyleGAN~\citep{stylegan} as $G$. PPA proceeds in three stages: (i) \emph{initial selection}, which samples random latent codes and retains those with the highest target-class confidence; (ii) \emph{optimization} of Eq.~\ref{eq:mia_gan}, which updates the selected latent codes using a Poincaré loss, with random image transformations; and (iii) \emph{final selection}, which keeps reconstructions with the highest target-class confidence under random augmentations. PPDG builds on this pipeline and additionally fine-tunes $G$ on pseudo-private data. Thus, target-model outputs guide all three stages: prediction scores guide both selection stages, while normalized logits provide the optimization signal.

\textbf{Inversion dynamics.} Let $x_t=G(z_t)$ be the reconstruction at step $t$. We track
\begin{equation}
\begin{aligned}
p_t &= [T_\theta(x_t)]_c,\\
d_t &= \min_{x\in\mathcal{D}_{\mathrm{priv}}^c}\|\phi_F(x_t)-\phi_F(x)\|_2,\\
g_t &= \|\nabla_{z_t}\mathcal{L}_{\mathrm{inv}}\|_2,
\end{aligned}
\label{eq:inversion_metrics}
\end{equation}
i.e., the target-class confidence, the distance to the nearest private sample of class $c$ in the feature space, and the gradient norm. Note that $d_t$ is used only for analysis and is unavailable to the attacker.


\subsection{Model Inversion Defenses}

Training-based defenses modify the training of $T_\theta$ to limit what Eq.~\ref{eq:mia_gan} can recover while preserving test accuracy. We study three representative defenses built on different principles. BiDO~\citep{bido} minimizes the dependency between inputs and intermediate representations while maximizing that between representations and labels. MIDRE~\citep{RE} applies random erasing to training images, so the model observes only partial private features. Neg.LS~\citep{ls} adopts negative label smoothing (\(\alpha<0\)), in contrast to conventional positive label smoothing (Pos.LS, \(\alpha>0\)), which has been shown to increase inversion leakage~\citep{ls}.

\subsection{Mean OOD Confidence and False Certainty}
\label{sec:ood_false_certainty}

Let $P_{\mathrm{ID}}$ denote the distribution of the private training data. Out-of-distribution (OOD) samples are natural images whose labels lie outside $\{1,\ldots,C\}$, \emph{e.g.}, faces of identities unseen during training; we denote their distribution by $P_{\mathrm{OOD}}$. We measure predictive confidence with the maximum softmax probability (MSP)~\citep{hendrycks2016baseline},
\begin{equation}
\begin{aligned}
m_\theta(x) &= \max_{k\in\{1,\ldots,C\}}P(k\mid x;T_\theta),\\
\mu_\theta(q) &= \mathbb{E}_{x\sim q}[m_\theta(x)],
\end{aligned}
\label{eq:ood_conf}
\end{equation}
where $\mu_\theta(q)$ is the average confidence over a distribution $q$. We denote the average MSP over OOD samples by $\mu_{\mathrm{OOD}}\equiv\mu_\theta(P_{\mathrm{OOD}})$. Unlike $p_t$, $m_\theta$ takes the maximum over all classes. A reliable model should satisfy $\mu_{\mathrm{OOD}}\ll\mu_\theta(P_{\mathrm{ID}})$, yet neural networks often produce confident predictions far from the training data~\citep{hein2019relu}. We say a model exhibits \emph{false certainty} when it assigns high confidence to OOD samples, i.e., high $\mu_{\mathrm{OOD}}$. FPR95 and AUROC~\citep{hendrycks2016baseline} are also reported.


\textbf{Relation to BiDO+.} BiDO+~\citep{bido+} observed OOD overconfidence in BiDO~\citep{bido} and mitigates it by additionally training on auxiliary OOD data, treating it as a calibration issue to be corrected on top of an existing defense. Our work differs in three aspects. \emph{(i) Scope:} we examine false certainty across defenses built on different principles, not only BiDO. \emph{(ii) Role:} rather than treating it as a side effect, we study how it relates to inversion resistance itself (Sec.~\ref{sec: rethinking}). \emph{(iii) Remedy:} instead of adding auxiliary OOD data, we show that a different training method, Input Mixup, attains inversion resistance without false certainty (Sec.~\ref{sec: ambiguity}).



\begin{figure*}[t]
    \centering

    \begin{subfigure}[b]{0.57\linewidth}
        \centering
        \includegraphics[width=\linewidth]{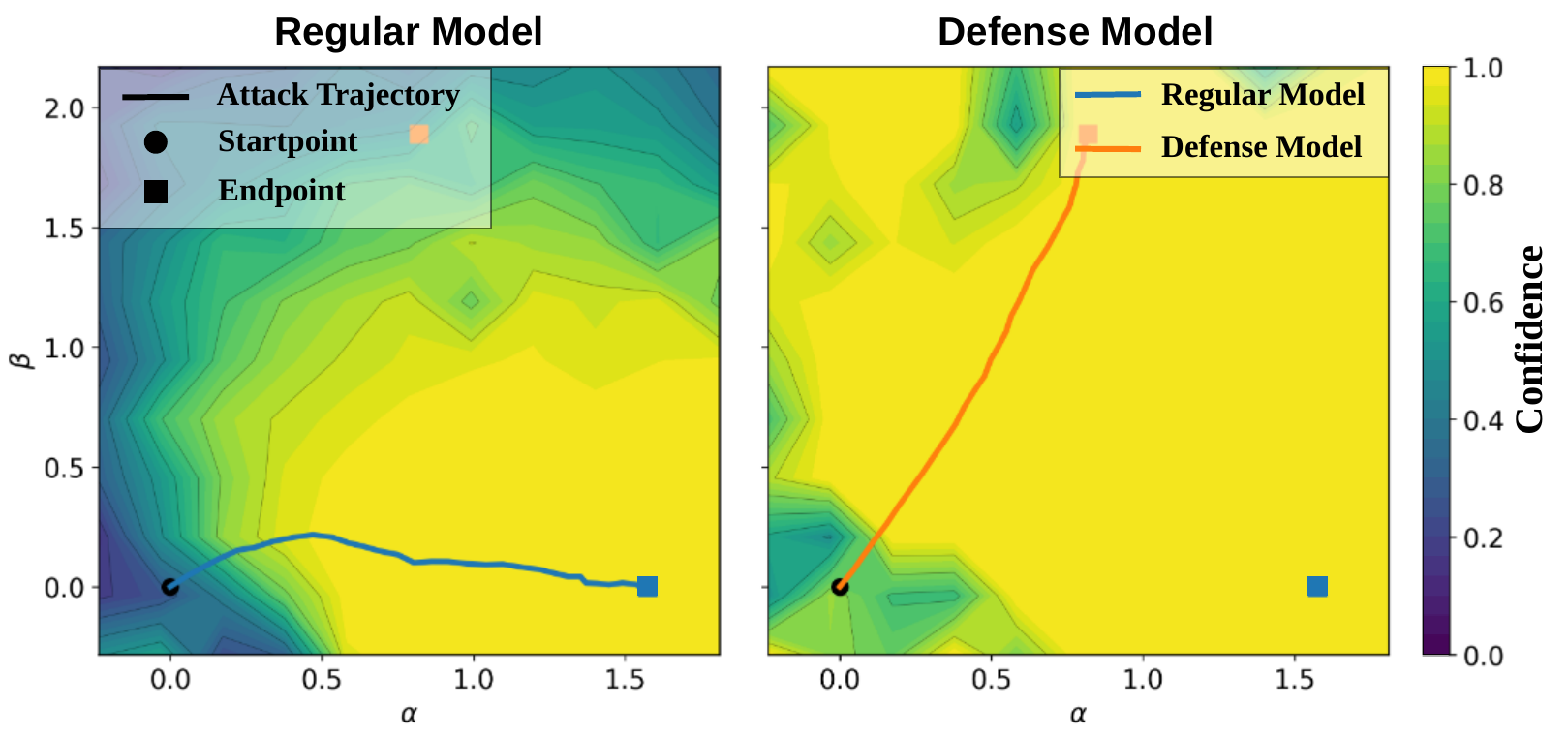}
        \caption{Confidence Landscapes}
        \label{fig:a.landscape}
    \end{subfigure}
    \hfill
    \begin{subfigure}[b]{0.415\linewidth}
        \centering
        \includegraphics[width=\linewidth]{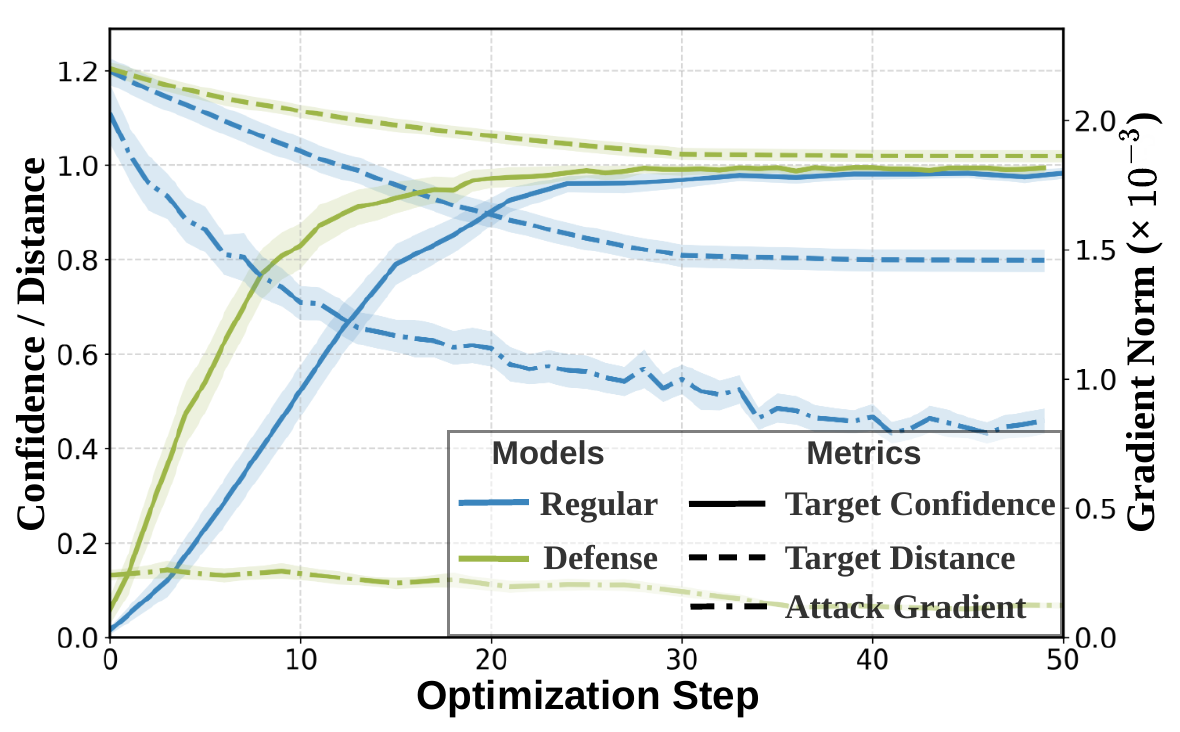}
        \caption{Inversion Dynamics}
        \label{fig:b.dynamics}
    \end{subfigure}

    \caption{\textbf{Confidence landscapes and inversion dynamics.} (a) Target-class confidence and inversion trajectories on a 2D latent-space slice. (b) Target-class confidence, target distance, and gradient norm over optimization steps.}
    \label{fig:traj&dynamics}
\end{figure*}

\begin{table}[t]
\centering
\small
\setlength{\tabcolsep}{3pt}
\renewcommand{\arraystretch}{1.18}
\caption{\textbf{OOD confidence and correlation analysis of MIA defenses.}
Pearson correlations measure the association between OOD confidence
and attack accuracy within each defense.}
\label{tab:ood_correlation}

\begin{tabular*}{\columnwidth}
{@{\extracolsep{\fill}}lccccc@{}}
\toprule
& \multicolumn{3}{c}{OOD Metrics}
& \multicolumn{2}{c}{Pearson} \\
\cmidrule(lr){2-4}
\cmidrule(lr){5-6}
\textbf{Method}
& $\mu_{\mathrm{OOD}}\downarrow$
& FPR95$\downarrow$
& AUROC$\uparrow$
& $r$
& $p$ \\
\midrule
Regular & 0.42 & 19.44 & 93.38 & -- & -- \\
\midrule
BiDO
& 0.60 & 23.00 & 94.54
& \textbf{-0.962}
& $1.38\times10^{-4}$ \\

MIDRE
& 0.88 & 27.80 & 89.54
& \textbf{-0.939}
& $5.34\times10^{-4}$ \\

Neg.LS
& 0.90 & 97.55 & 93.68
& \textbf{-0.908}
& $1.83\times10^{-3}$ \\
\bottomrule
\end{tabular*}
\label{tab:correlation}
\end{table}


\section{Rethinking MIA Defenses through OOD Overconfidence}
\label{sec: rethinking}

Existing MIA defenses employ diverse strategies to mitigate inversion leakage, yet the mechanisms underlying their effectiveness remain incompletely understood. OOD overconfidence has previously been reported for BiDO~\citep{bido,bido+}, raising the question of whether similar behavior occurs in other defenses and how it relates to inversion resistance. In this section, we investigate this relationship and examine its implications for inversion optimization.

\subsection{OOD Overconfidence and Inversion Resistance}
\label{sec3.1}

Tab.~\ref{tab:correlation} compares the OOD behavior of BiDO, MIDRE, and Neg.LS with that of the regular model. All three defenses assign markedly higher confidence to OOD samples: $\mu_{\mathrm{OOD}}$ rises from 0.42 for the regular model to 0.60, 0.88, and 0.90, respectively. Their FPR95 increases from 19.44\% to 23.00-97.55\%, while AUROC ranges from 89.54\% to 94.54\%, compared with 93.38\% for the regular model. Notably, Neg.LS exhibits particularly severe OOD overconfidence, with \(\mu_{\mathrm{OOD}}\) reaching 0.90 or even higher and FPR95 approaching 100\%, indicating that nearly all OOD samples are incorrectly accepted at the 95\% ID-recall operating point. Although these defenses rely on different principles, namely dependency regularization, data augmentation, and label smoothing, they share this behavior, indicating that false certainty is not specific to BiDO~\citep{bido+}.


False certainty is shared across these defenses, but whether its magnitude tracks inversion resistance within each defense remains unclear. We train $8$ configurations per method with varying hyperparameters at comparable test accuracy. Across all three defenses, higher $\mu_{\mathrm{OOD}}$ corresponds to lower attack accuracy, indicating stronger resistance. Pearson correlations are $-0.908$, $-0.939$, and $-0.962$ for Neg.LS, MIDRE, and BiDO, respectively, all with $p$-values on the order of $10^{-3}$ to $10^{-4}$ (Tab.~\ref{tab:correlation}).

However, we do not observe the same pattern across other evaluated defenses. For instance, TL~\citep{TL} and Trap~\citep{Trap-MID} achieve strong inversion resistance without exhibiting elevated OOD confidence (see Tab.~\ref{tab:main} for further comparisons). This suggests that the negative association between OOD confidence and attack accuracy is not universal across all existing MIA defenses, and that OOD overconfidence represents one possible route to inversion resistance rather than a necessary condition.


\subsection{Premature Saturation in Inversion}
\label{sec3.2}

Sec.~\ref{sec3.1} establishes that false certainty accompanies inversion resistance. We now show how it interferes with the attack. Recall that generative MIAs initialize from samples drawn from a learned prior, typically trained on public data with identities disjoint from the private training classes~\citep{ppdg}. These initial samples are therefore generally OOD with respect to the target model. The attack then optimizes their latent codes to increase the target-class prediction score, gradually seeking reconstructions of the target identity.

\textbf{Confidence landscapes.} Fig.~\ref{fig:a.landscape} visualizes target-class confidence over a 2D slice of the generator's latent space, computed on identical generated images for the regular model and the defended models discussed above. For the regular model, high confidence is confined to a small region around the target identity, while the confidence elsewhere remains relatively low. The landscape therefore acts as a reliable guide, where increasing confidence guides the attack toward the target. In contrast, the defended model assigns high confidence across much broader regions, including those corresponding to non-target identities. Crucially, the optimization trajectory terminates in one of these non-target regions rather than near the target identity. Consequently, the reconstruction achieves high target-class confidence without recovering the target, illustrating how false certainty misleads the attack into converging to an incorrect solution.

\textbf{Inversion dynamics.} Fig.~\ref{fig:b.dynamics} tracks the same attack step by step. Against the regular model, the target-class confidence $p_t$ rises gradually while the distance to the target $d_t$ steadily decreases, so each gain in confidence corresponds to progress toward the target. Against the defended model, $p_t$ saturates much earlier, while $d_t$ barely decreases. Once confidence saturates, the attack objective is essentially satisfied, and the gradient norm $g_t$ decays, leaving the attack little signal to refine the reconstruction. The attack thus ends far from the target despite reaching near-maximal confidence. We refer to this as \emph{premature saturation}.

These observations explain how false certainty yields inversion resistance: by assigning high target-class confidence to faces of other identities, the defended model satisfies the attack objective before the reconstruction reaches the target, thereby further demonstrating its effectiveness in limiting successful target reconstruction.

\begin{figure}[t]
    \centering
    \includegraphics[width=0.99\linewidth]{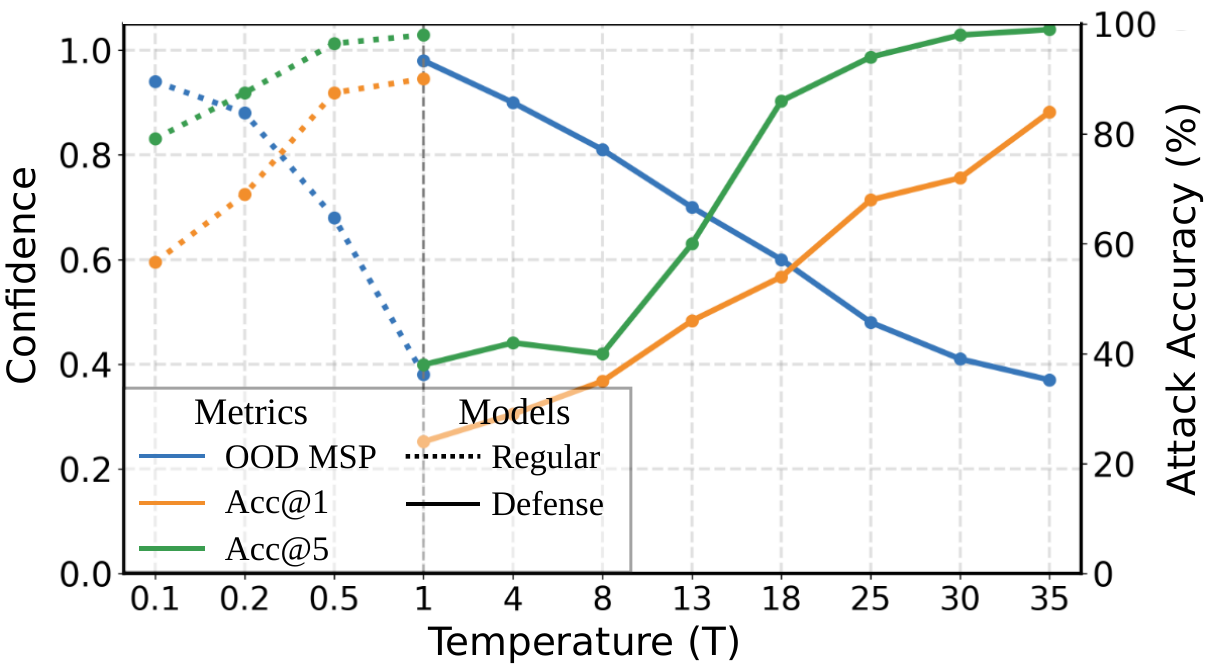}
    \caption{\textbf{Temperature scaling of defended and regular models.} OOD confidence and attack accuracy under different temperatures.}
    \label{fig:TS}
\end{figure}

\begin{figure}
    \centering
    \includegraphics[width=0.95\linewidth]{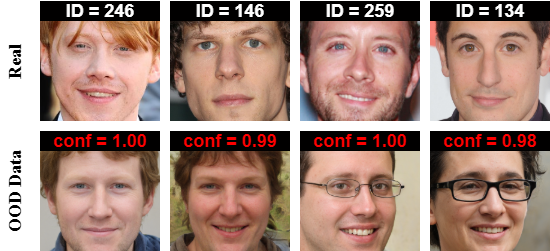}
    \caption{\textbf{OOD Overconfidence in Identity Classification.} Top: real samples from different identities. Bottom: OOD inputs assigned high prediction confidence by the defense model.}
    \label{fig:real vs ood}
\end{figure}


\subsection{Intervening on Confidence}
\label{sec3.3}

Secs.~\ref{sec3.1} and~\ref{sec3.2} establish an association between false certainty and inversion resistance. We now further investigate this relationship through a controlled confidence intervention using temperature scaling (TS)~\citep{ts}. TS rescales the model's logits before softmax: a temperature $T>1$ softens the output distribution and reduces prediction confidence on OOD data, whereas $0<T<1$ sharpens it. Importantly, this transformation preserves the predicted labels and thus leaves test accuracy unchanged. We apply TS to the target models to alter their OOD confidence, then perform inversion attacks against the resulting models to examine how attack performance varies with the induced confidence changes.

We first soften the three defended models using $T>1$. As $T$ increases from 4 to 35, $\mu_{\mathrm{OOD}}$ decreases from 0.90 to 0.37, while attack accuracy rises from 29\% to 84\% (top-1 attack accuracy) and from 38\% to 99\% (top-5 attack accuracy), with test accuracy remaining around 94\%. Notably, although $\mu_{\mathrm{OOD}}$ at $T=35$ falls below that of the regular model (0.38), attack accuracy (84\%) remains below its original level (90\%). Conversely, sharpening the regular model using $T<1$ increases its $\mu_{\mathrm{OOD}}$ from 0.38 to 0.94, while PPA attack accuracy decreases from 90\% to 56\%, despite unchanged predicted labels and test accuracy (Fig.~\ref{fig:TS}). These results show that changes in OOD confidence are accompanied by corresponding changes in inversion performance in both directions, supporting OOD overconfidence as a contributing factor to inversion resistance.

Nevertheless, attack performance does not fully recover even when the defended models' $\mu_{\mathrm{OOD}}$ falls below that of the regular model, suggesting that OOD overconfidence is not the sole factor underlying inversion resistance. Other mechanisms may also contribute. For example, \citet{AT-AT} link MIA defenses to a privacy-robustness trade-off arising from increased reliance on non-robust features, including in BiDO. This suggests that factors beyond OOD overconfidence may also play a role in inversion resistance.


\subsection{Defense at the Cost of False Certainty}

\begin{figure*}[t]
    \centering

    \begin{subfigure}[b]{0.63\linewidth}
        \centering
        \includegraphics[width=\linewidth]{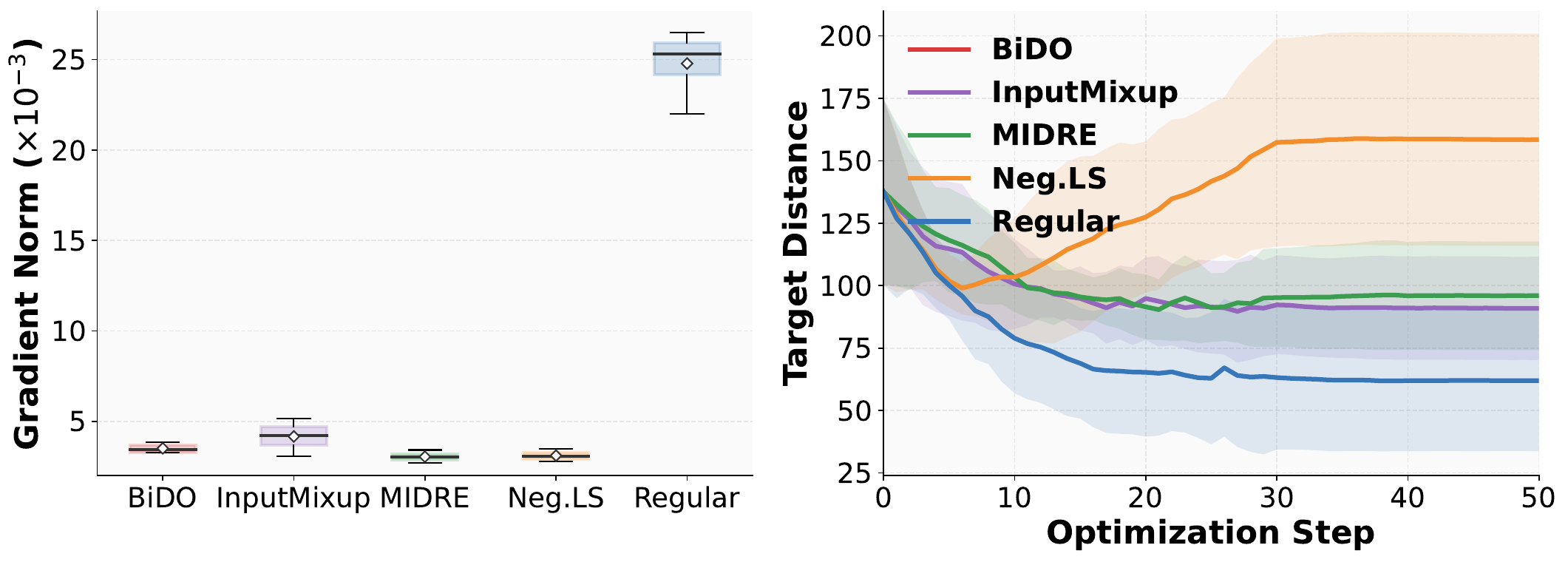}
        \caption{Attack optimization dynamics under Mixup.}
        \label{fig:a.gradient_norm_and_target_distance}
    \end{subfigure}
    \hfill
    \begin{subfigure}[b]{0.36\linewidth}
        \centering
        \includegraphics[width=\linewidth]{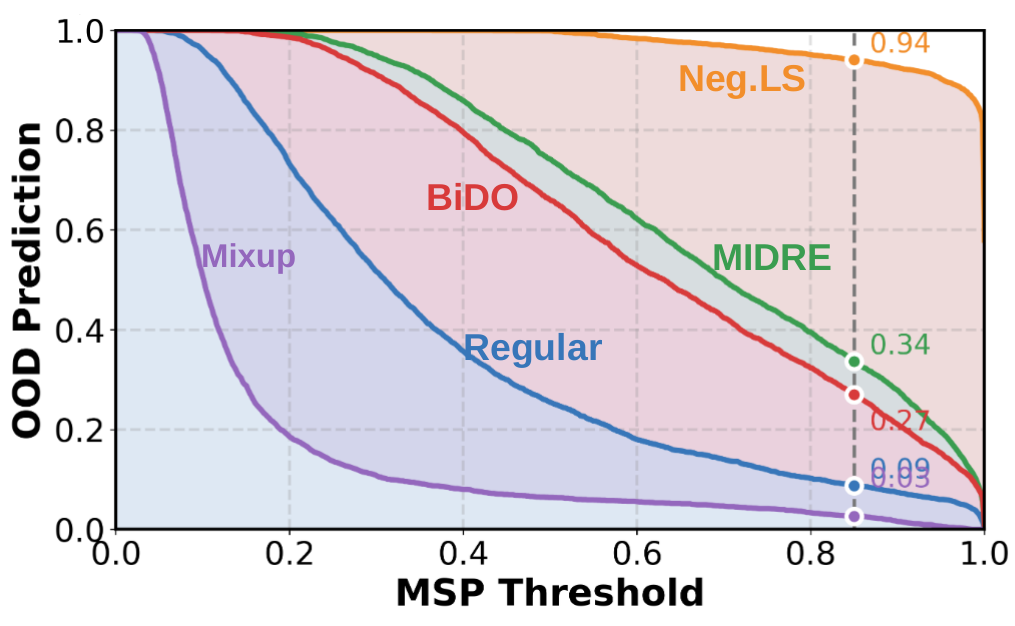}
        \caption{OOD acceptance across confidence thresholds.}
        \label{fig:b.ood_msp}
    \end{subfigure}

    \caption{\textbf{Mixup's inversion dynamics (a) and OOD acceptance across confidence thresholds (b).}}
    \label{fig:gradient_norm_and_msp}
\end{figure*}


The resistance gained through false certainty comes at a cost: the defended models become unreliable on inputs outside their training identities. In open-set face recognition, probe images may come from individuals absent from the enrolled gallery and should be rejected rather than matched to a known identity~\citep{vareto2024open}. Such rejection typically relies on thresholding the model's confidence~\citep{hendrycks2016baseline}, which false certainty undermines. Fig.~\ref{fig:real vs ood} illustrates this consequence, showing that the defended models assign near-certain confidence to known identities when presented with faces of unseen individuals. In watchlist applications, such errors lead to false alarms and misidentifications, and similar risks arise in other safety-critical domains such as medical diagnosis~\citep{weng2025first} and autonomous driving~\citep{araujo2024road}.

These observations expose a trade-off: the OOD overconfidence contributing to inversion resistance also makes defended models unreliable on unfamiliar inputs. This raises a natural question: can strong inversion resistance be achieved without such false certainty?

\section{Decoupling Inversion Resistance from False Certainty}
\label{sec: ambiguity}


Sec.~\ref{sec: rethinking} identifies OOD overconfidence as one effective route to inversion resistance, but this protection comes at the cost of unreliable predictions on unfamiliar inputs. This raises an important question: can strong inversion resistance be achieved without introducing such a reliability cost? We investigate this question through Input Mixup (hereafter Mixup)~\citep{mixup}, a standard training technique that interpolates both training inputs and their labels. As shown in Tab.~\ref{tab:overall_comparison}, Mixup achieves the lowest attack accuracy in 7 of the 8 evaluated settings, while keeping OOD confidence at or below that of the regular model (Tab.~\ref{tab:main}). Unlike BiDO+~\citep{bido+}, Mixup requires no auxiliary OOD data. It retains the standard cross-entropy loss on interpolated samples, without introducing additional loss terms or architectural modifications. These results demonstrate that strong inversion resistance can be achieved without exhibiting the elevated OOD confidence observed in the defenses of Sec.~\ref{sec: rethinking}. We further investigate how input interpolation contributes to this effect.


\subsection{Inversion Resistance without False Certainty}
\label{sec:mixup_false_certainty}

Input Mixup~\citep{mixup} trains the model on convex combinations of training pairs and their labels:
\begin{equation}
\begin{aligned}
\tilde{\mathbf{x}} &= \lambda\mathbf{x}_i+(1-\lambda)\mathbf{x}_j,\\
\tilde{\mathbf{y}} &= \lambda\mathbf{y}_i+(1-\lambda)\mathbf{y}_j,
\end{aligned}
\label{eq:input_mixup}
\end{equation}
where $\lambda\sim\operatorname{Beta}(\alpha,\alpha)$. By supervising the model on interpolated inputs and their corresponding soft labels, Mixup encourages predictions that better reflect the mixing proportions, rather than assigning intermediate inputs confidently to a single identity (Tab.~\ref{tab:mixup_analysis}). Mixup is also known to improve predictive calibration~\citep{onmixuptrain}.

\textbf{Mixup resists inversion without elevated OOD confidence.} Despite its strong inversion resistance, Mixup does not exhibit the OOD overconfidence observed in Sec.~\ref{sec: rethinking}. On FaceScrub, its $\mu_{\mathrm{OOD}}$ is 0.27, substantially below that of the regular model (0.42) and the three defenses studied in Sec.~\ref{sec: rethinking} (0.60--0.90). This pattern also holds on CelebA, where Mixup achieves a lower $\mu_{\mathrm{OOD}}$ than the regular model (0.41 vs.\ 0.45; Tab.~\ref{tab:main}). Fig.~\ref{fig:b.ood_msp} further shows that Mixup rejects a larger proportion of OOD samples across confidence thresholds. Thus, unlike the defenses in Sec.~\ref{sec: rethinking}, Mixup achieves strong inversion resistance without increasing confidence on unseen identities.

\begin{table}[t]
\centering
\small
\setlength{\tabcolsep}{4pt}
\renewcommand{\arraystretch}{1.12}
\setlength{\aboverulesep}{0pt}
\setlength{\belowrulesep}{0pt}
\caption{\textbf{Effects of Mixup and Pos.LS on MIA vulnerability and identity representations.}}
\label{tab:mixup_analysis}
\begin{tabular}{l|ccccc}
\toprule
\rowcolor{tblHeader}
\textbf{Method}
& \textbf{PPA$\downarrow$}
& \textbf{Intra$\uparrow$}
& \textbf{Inter$\downarrow$}
& \textbf{KL$\downarrow$}
& \textbf{Pair Mass$\uparrow$} \\
\midrule
\textbf{Regular}
& 85.93 & 0.10 & 0.48 & 2.23 & 0.68 \\

\textbf{Pos.LS}
& 86.27 & 0.06 & 0.93 & 0.84 & 0.44 \\

\midrule
\rowcolor{tblHighlight}
\textbf{Mixup}
& \textbf{10.93} & \textbf{0.20} & \textbf{0.37} & \textbf{0.61} & \textbf{0.71} \\
\bottomrule
\end{tabular}%
\end{table}

Interestingly, Fig.~\ref{fig:a.gradient_norm_and_target_distance} shows that Mixup hinders inversion optimization, yielding larger target distances and smaller gradient norms than the regular model. Although this behavior resembles that of the defenses in Sec.~\ref{sec: rethinking}, their underlying confidence behavior differs. While these defenses induce premature saturation through elevated confidence on non-target inputs, Mixup impedes optimization without such confidence inflation. Thus, similar inversion resistance can arise through different routes, and OOD overconfidence is not necessary for impeding inversion. Notably, Neg.LS exhibits the largest target distance in Fig.~\ref{fig:a.gradient_norm_and_target_distance}, consistent with its particularly strong OOD overconfidence (Tab.~\ref{tab:ood_correlation}) and the resulting premature saturation of attack optimization. Overall, these findings reveal distinct optimization mechanisms underlying inversion resistance, with Mixup providing an alternative that does not rely on OOD overconfidence.



\begin{table*}[t]
\centering
\footnotesize
\setlength{\tabcolsep}{2.5pt} 
\renewcommand{\arraystretch}{1.0}
\setlength{\aboverulesep}{0pt}
\setlength{\belowrulesep}{0pt}
\caption{\textbf{Comparison of MIA defenses across different architectures and datasets, evaluated with PPA and PPDG.}}
\label{tab:overall_comparison}
\resizebox{\textwidth}{!}{%
\begin{tabular}{l|ccc|ccc|ccc|ccc}
\toprule
\rowcolor{tblHeader}
& \multicolumn{6}{>{\columncolor{tblHeader}}c|}{\textbf{FaceScrub}}
& \multicolumn{6}{>{\columncolor{tblHeader}}c}{\textbf{CelebA}} \\
\cline{2-7}\cline{8-13}
\rowcolor{tblHeader}
& \multicolumn{3}{>{\columncolor{tblHeader}}c|}{\textbf{RN-152}}
& \multicolumn{3}{>{\columncolor{tblHeader}}c|}{\textbf{DN-121}}
& \multicolumn{3}{>{\columncolor{tblHeader}}c|}{\textbf{RN-152}}
& \multicolumn{3}{>{\columncolor{tblHeader}}c}{\textbf{DN-121}} \\
\cline{2-4}\cline{5-7}\cline{8-10}\cline{11-13}
\rowcolor{tblHeader}
\multirow{-3}{*}{\textbf{Method}} & \textbf{Test$\uparrow$}
& \textbf{PPA$\downarrow$}
& \textbf{PPDG$\downarrow$}
& \textbf{Test$\uparrow$}
& \textbf{PPA$\downarrow$}
& \textbf{PPDG$\downarrow$}
& \textbf{Test$\uparrow$}
& \textbf{PPA$\downarrow$}
& \textbf{PPDG$\downarrow$}
& \textbf{Test$\uparrow$}
& \textbf{PPA$\downarrow$}
& \textbf{PPDG$\downarrow$} \\
\midrule
\textbf{Regular}
& 96.39& 85.93 & 87.27& 97.11& 90.06& 95.73& 91.78& 73.26& 78.37& 89.91& 65.07& 70.56\\
\multicolumn{13}{>{\columncolor{tblSection}}l}{\textbf{\textit{Defenses with Identified OOD Overconfidence}}} \\
\textbf{BiDO}
& 92.54 \bad{-3.85} & 49.40 \good{-36.53} & 50.90 \good{-36.37} 
& 92.23& 57.73& 60.56 
& 81.82 \bad{-9.96} & 19.20 \good{-54.06} & 25.75 \good{-52.62} 
& 80.18& 49.36& 61.28\\
\textbf{MIDRE}
& 90.60 \bad{-5.79} & 23.33 \good{-62.60} & 24.26 \good{-63.01} 
& 89.64& 39.61& 38.96 
& 76.80 \bad{-14.98} & 34.80 \good{-38.46} & 36.12 \good{-42.25} 
& 84.67& 31.34& 40.47\\
\textbf{Neg.LS}
& 92.21 \bad{-4.18} & 16.00 \good{-69.93} & 19.37 \good{-67.90} 
& 93.42& 30.69& 29.18 
& 86.92 \bad{-4.86} & 37.28 \good{-35.98} & 38.98 \good{-39.39} 
& 87.90& 45.63& 48.71\\
\multicolumn{13}{>{\columncolor{tblSection}}l}{\textbf{\textit{Other Defense Baselines}}} \\
\textbf{TL}
& 91.71 \bad{-4.68} & 25.89 \good{-60.04} & 30.08 \good{-57.19} 
& 91.44& 25.88& 30.78 
& 75.51 \bad{-16.27} & 24.52 \good{-48.74} & 32.07 \good{-46.30} 
& 79.45& 29.37& 36.42\\
\textbf{Trap}
& 93.10 \bad{-3.29} & 70.39 \good{-15.54} & 75.98 \good{-11.29} 
& 93.65& 56.70& 58.73 
& 87.26 \bad{-4.52} & 53.92 \good{-19.34} & 58.47 \good{-19.90} 
& 87.89& 37.64& 36.17\\
\textbf{LoFT}
& 94.79 \bad{-1.60} & 25.07 \good{-60.86} & 24.47 \good{-62.80} 
& 92.13& 35.08& 36.33 
& 75.40 \bad{-16.38} & 20.40 \good{-52.86} & 24.33 \good{-54.04} 
& 75.88& \textbf{21.36}& 32.06\\
\multicolumn{13}{>{\columncolor{tblSection}}l}{\textbf{\textit{OOD Overconfidence-Aware Defense}}} \\
\textbf{BiDO+}
& 93.21 \bad{-3.18} & 52.39 \good{-33.54} & 54.67 \good{-32.60} 
& 90.01& 45.77& 48.60 
& 85.83 \bad{-5.95} & 36.21 \good{-37.05} & 36.43 \good{-41.94} 
& 82.19& 43.98& 45.02\\
\midrule
\rowcolor{tblHighlight}
\textbf{Mixup}
& 94.64 \bad{-1.75} & \textbf{10.93} \good{-75.00} & \textbf{12.34} \good{-74.93} 
& 97.75& \textbf{17.20}& \textbf{15.56} 
& 87.67 \bad{-4.11} & \textbf{19.01} \good{-54.25} & \textbf{21.04} \good{-57.33} 
& 89.34& 27.20& \textbf{27.89}\\
\bottomrule
\end{tabular}%
}
\end{table*}



\begin{table*}[t]
\centering
\footnotesize
\setlength{\tabcolsep}{3pt}
\renewcommand{\arraystretch}{1.0}
\setlength{\aboverulesep}{0pt}
\setlength{\belowrulesep}{0pt}
\caption{\textbf{Comparison of MIA defenses against PPA for ResNet-152 trained on FaceScrub and CelebA.}}
\label{tab:main}
\resizebox{\textwidth}{!}{%
\begin{tabular}{l|ccccc|ccccc}
\toprule
\rowcolor{tblHeader}
& \multicolumn{5}{>{\columncolor{tblHeader}}c|}{\textbf{FaceScrub}} & \multicolumn{5}{>{\columncolor{tblHeader}}c}{\textbf{CelebA}} \\
\cline{2-6}\cline{7-11}
\rowcolor{tblHeader}
\multirow{-2}{*}{\textbf{Model}} & Test$\uparrow$& $\mu_{\mathrm{OOD}}\downarrow$ & PPA$\downarrow$& $\delta_{face}\uparrow$ & $\delta_{eval}\uparrow$
& Test$\uparrow$& $\mu_{\mathrm{OOD}}\downarrow$ & PPA$\downarrow$& $\delta_{face}\uparrow$ & $\delta_{eval}\uparrow$ \\
\midrule
\textbf{Regular} & 96.39 & 0.42 & 85.93\err{1.51} & 0.78\err{0.03} & 96.33\err{7.04} & 91.78 & 0.45 & 73.26\err{2.41} & 0.79\err{0.04} & 278.39\err{13.31} \\
\midrule
\textbf{BiDO} & 92.54 & 0.60 & 49.40\err{0.98} & 0.92\err{0.03} & 132.14\err{12.34} & 81.82 & 0.62 & 19.20\err{2.13} & 0.95\err{0.03} & 329.83\err{14.28} \\
\textbf{MIDRE} & 90.60 & 0.88 & 23.33\err{1.57} & 0.98\err{0.04} & 150.42\err{14.96} & 76.80 & 0.92 & 34.80\err{1.43} & 1.37\err{0.02} & 305.75\err{9.56} \\
\textbf{Neg.LS} & 92.21 & 0.90 & 16.00\err{1.81} & 1.17\err{0.06} & 203.60\err{15.89} & 86.92 & 0.84 & 37.28\err{3.01} & 0.94\err{0.03} & 322.94\err{10.23} \\
\midrule
\textbf{TL} & 91.71 & 0.48 & 25.89\err{1.21} & 1.04\err{0.03} & 147.57\err{10.34} & 75.51 & 0.52 & 24.52\err{2.36} & 1.12\err{0.04} & 339.09\err{12.36} \\
\textbf{Trap} & 93.10 & 0.35 & 70.39\err{2.67} & 0.74\err{0.02} & 114.21\err{12.32} & 87.26 & 0.54 & 53.92\err{4.37} & 0.93\err{0.03} & 275.06\err{17.72} \\
\textbf{LoFT} & 94.79 & 0.54 & 25.07\err{1.65} & 1.06\err{0.04} & 137.63\err{13.98} & 75.40 & 0.54 & 20.40\err{6.53} & 1.37\err{0.05} & 412.31\err{20.03} \\
\midrule
\textbf{BiDO+} & 93.21 & 0.22 & 52.39\err{1.11} & 0.97\err{0.02} & 134.53\err{9.90} & 85.83 & 0.24 & 36.21\err{2.20} & 1.08\err{0.04} & 315.66\err{14.40} \\
\midrule
\rowcolor{tblHighlight}
\textbf{Mixup} & 94.64 & 0.27 & 10.93\err{1.90} & 1.16\err{0.05} & 163.47\err{13.67} & 87.67 & 0.41 & 19.01\err{2.30} & 1.00\err{0.03} & 332.90\err{10.67} \\
\bottomrule
\end{tabular}%
}
\end{table*}

\subsection{Interpolation, Not Soft Labels}
\label{sec:mixup_interpolation}

Mixup interpolates both inputs and labels, so its resistance might simply come from training on soft targets. We test this with positive label smoothing (Pos.LS), which also trains on soft targets but does not interpolate inputs. Pos.LS provides no protection: its attack accuracy (86.27\%) matches that of the regular model (85.93\%), consistent with prior work~\citep{ls}, whereas Mixup reduces it to 10.93\% (Tab.~\ref{tab:mixup_analysis}). This comparison suggests that soft-label supervision alone is insufficient to account for Mixup's resistance, highlighting input interpolation as a key ingredient.

Tab.~\ref{tab:mixup_analysis} further characterizes how different models shape inter-identity representations. In feature space, cosine distances between penultimate-layer features show that Pos.LS produces tighter, more separated identity clusters than the regular model (intra-class distance 0.06 vs.\ 0.10; inter-class distance 0.93 vs.\ 0.48), whereas Mixup produces looser, closer clusters (0.20 and 0.37). In prediction space, we evaluate interpolated inputs using \textbf{Pair Mass}, the probability assigned to the source identities, and the \textbf{KL divergence} between their relative probabilities and the mixing ratio. The regular model retains most probability mass on the source identities (0.68) but poorly follows the mixing ratio (KL 2.23), whereas Pos.LS follows it more closely (0.84) but spreads mass across other identities (0.44). Mixup alone achieves both (0.71 and 0.61).

These results suggest that input interpolation changes how the model represents and predicts samples between identities. Unlike Pos.LS, which produces tighter, more separated identity clusters, Mixup yields more dispersed representations while keeping prediction mass concentrated on source identities and reflecting their mixing proportions. This distinction matters for inversion: rather than producing sharply separated responses, Mixup encourages predictions that reflect interpolation between identities. Together with weaker attack gradients in Fig.~\ref{fig:a.gradient_norm_and_target_distance}, these findings suggest that input interpolation reshapes the identity-specific signals exploited by inversion attacks, potentially hindering target reconstruction without elevated OOD confidence.

\begin{table*}[t]
\centering
\footnotesize
\setlength{\tabcolsep}{4pt}
\renewcommand{\arraystretch}{1.0}
\setlength{\aboverulesep}{0pt}
\setlength{\belowrulesep}{0pt}
\caption{\textbf{Privacy-robustness efficiency of Mixup and AT-AT across datasets and architectures.}}
\label{tab:adversarial_robustness}
\newcommand{\tabfivehl}[1]{\cellcolor{tblHighlight}#1}
\resizebox{\textwidth}{!}{%
\begin{tabular}{ll|l|ccc|ccc|cc}
\toprule
\rowcolor{tblHeader}
& & & & & & \multicolumn{3}{>{\columncolor{tblHeader}}c|}{\textbf{Robust Acc$\uparrow$}}
& \multicolumn{2}{>{\columncolor{tblHeader}}c}{\textbf{Cost$\downarrow$}}\\
\cline{7-9}\cline{10-11}
\rowcolor{tblHeader}
\multirow{-2}{*}{\textbf{Dataset}} & \multirow{-2}{*}{\textbf{Model}} & \multirow{-2}{*}{\textbf{Method}}
& \multirow{-2}{*}{\textbf{Test Acc$\uparrow$}} & \multirow{-2}{*}{\textbf{PPA$\downarrow$}} & \multirow{-2}{*}{\textbf{PPDG$\downarrow$}}
& $\epsilon=0.0025$ & $\epsilon=0.0005$ & $\epsilon=10^{-5}$
& $\epsilon=0.0005$ & $\epsilon=10^{-5}$\\
\midrule
& & Regular & 96.39 & 85.93 & 87.27 & 0.056 & 0.847 & 0.941 & \textemdash & \textemdash\\
& & AT-AT & 92.10 & 20.69 & 22.37 & 0.000 & 0.000 & 0.304 & 1.30 & 0.98\\
& \multirow{-3}{*}{RN-152} & \tabfivehl{\textbf{Input Mixup}} & \tabfivehl{94.64} & \tabfivehl{\textbf{10.93}} & \tabfivehl{\textbf{12.34}} & \tabfivehl{0.000} & \tabfivehl{0.756} & \tabfivehl{0.923} & \tabfivehl{\textbf{0.12}} & \tabfivehl{\textbf{0.02}}\\
& & Regular & 97.11 & 90.06 & 95.73 & 0.260 & 0.945 & 0.989 & \textemdash & \textemdash\\
& & AT-AT & 92.36 & 25.76 & 24.64 & 0.000 & 0.000 & 0.590 & 1.47 & 0.62\\
\multirow{-6}{*}{\textit{FaceScrub}} & \multirow{-3}{*}{DN-121} & \tabfivehl{\textbf{Input Mixup}} & \tabfivehl{97.75} & \tabfivehl{\textbf{17.20}} & \tabfivehl{\textbf{15.56}} & \tabfivehl{0.000} & \tabfivehl{0.858} & \tabfivehl{0.982} & \tabfivehl{\textbf{0.12}} & \tabfivehl{\textbf{0.01}}\\
\midrule
& & Regular & 91.78 & 73.26 & 78.37 & 0.037 & 0.707 & 0.793 & \textemdash & \textemdash\\
& & AT-AT & 84.33 & 33.56 & 36.55 & 0.000 & 0.000 & 0.548 & 1.78 & 0.62\\
& \multirow{-3}{*}{RN-152} & \tabfivehl{\textbf{Input Mixup}} & \tabfivehl{87.67} & \tabfivehl{\textbf{19.01}} & \tabfivehl{\textbf{21.04}} & \tabfivehl{0.000} & \tabfivehl{0.589} & \tabfivehl{0.760} & \tabfivehl{\textbf{0.22}} & \tabfivehl{\textbf{0.06}}\\
& & Regular & 89.91 & 65.07 & 70.56 & 0.187 & 0.843 & 0.901 & \textemdash & \textemdash\\
& & AT-AT & 81.67 & 56.37 & 58.66 & 0.000 & 0.000 & 0.876 & \textemdash & \textemdash\\
\multirow{-6}{*}{\textit{CelebA}} & \multirow{-3}{*}{DN-121} & \tabfivehl{\textbf{Input Mixup}} & \tabfivehl{89.34} & \tabfivehl{\textbf{27.20}} & \tabfivehl{\textbf{27.89}} & \tabfivehl{0.000} & \tabfivehl{0.737} & \tabfivehl{0.893} & \tabfivehl{\textbf{0.28}} & \tabfivehl{\textbf{0.02}}\\
\bottomrule
\end{tabular}%
}
\parbox{\textwidth}{\vspace{3pt}\footnotesize
A dash marks the Regular reference or a PPA reduction below 10 percentage points (CelebA/DN-121/AT-AT).}
\end{table*}


\section{Experiments}
\label{sec: exp}

\subsection{Experimental Setting}

\textbf{Datasets and target models.} Following prior studies~\citep{manifold,PPA,ls}, we use FaceScrub~\citep{facescrub} (530 identities) and the 1,000 most frequent identities of CelebA~\citep{celeba}, with images center-cropped and resized to $224\times224$. Target models are ResNet-152~\citep{dnn2} and DenseNet-121~\citep{densenet}, initialized with ImageNet-pretrained weights~\citep{imagenet} and fine-tuned on each dataset.

\textbf{OOD data.} Following BiDO+~\citep{bido+}, we use domain-matched OOD data: face images of CelebA identities disjoint from the training classes of each dataset. 

\textbf{Attacks.} Following prior defenses~\citep{TL,bido,ls,RE,loft}, we evaluate against two white-box attacks, PPA~\citep{PPA} and PPDG~\citep{ppdg}, under their original protocols, including the default Poincar\'e identity loss. We focus on white-box attacks because our analysis concerns the confidence landscapes and gradient-based optimization dynamics underlying inversion resistance.

\textbf{Metrics.} We measure attack success using Top-1 (Acc@1) and Top-5 (Acc@5) accuracy, evaluated by an independent Inception-v3 model~\citep{inception} on reconstructed images. Reconstruction quality is further assessed through feature distances to private images, measured using FaceNet~\citep{facenet} ($\delta_{\mathrm{face}}$) and the evaluation model ($\delta_{\mathrm{eval}}$) respectively, where larger distances indicate less faithful reconstructions. We evaluate OOD behavior using $\mu_{\mathrm{OOD}}$, FPR95, and AUROC (Sec.~\ref{sec:ood_false_certainty}). All results are reported as mean $\pm$ standard deviation over 3 runs.

\textbf{Baselines.} Besides BiDO~\citep{bido}, MIDRE~\citep{RE}, and Neg.LS~\citep{ls}, we compare with TL-DMI (TL)~\citep{TL}, Trap-MID (Trap)~\citep{Trap-MID}, and LoFT~\citep{loft}, which follow other defense strategies, and with BiDO+~\citep{bido+}, which explicitly addresses OOD overconfidence. For the robustness analysis (Sec.~\ref{sec:adversarial_robustness}), we additionally include AT-AT~\citep{AT-AT}. Baseline hyperparameters follow the original papers. For Input Mixup, we select $\alpha=2.5$ on FaceScrub/ResNet-152 and reuse it without retuning for all other datasets, architectures, and attacks; it is not the best value in every setting. All other training settings match the regular model.


\subsection{Main Results} 
\label{sec:main_results}

Tab.~\ref{tab:overall_comparison} compares 7 defenses and Mixup across 2 datasets $\times$ 2 architectures $\times$ 2 attacks. Mixup achieves the lowest attack accuracy in 7 of the 8 settings under the evaluated attack protocols, without retuning. For ResNet-152 under PPA, Mixup also achieves lower reported attack accuracy than every baseline on both datasets at the tested strengths $\alpha=2.0$ and $2.5$. This resistance comes with little loss in utility: Mixup's test accuracy drops by at most 4.11\% and even improves on FaceScrub with DenseNet-121. Baselines with comparable resistance pay considerably more in terms of test accuracy. On CelebA with ResNet-152, for example, TL~\citep{TL} and LoFT~\citep{loft} reach attack accuracies similar to Mixup's but lose over 16\% of test accuracy relative to the regular model.

Tab.~\ref{tab:main} adds OOD confidence and reconstruction distances for ResNet-152 under PPA. The defenses identified in Sec.~\ref{sec: rethinking} pair low attack accuracy with high $\mu_{\mathrm{OOD}}$ (0.60--0.92). Among the strongest defenses, Mixup is the only one whose $\mu_{\mathrm{OOD}}$ is below the regular model's on both datasets. BiDO+ provides further evidence that OOD overconfidence is not the sole source of inversion resistance. By introducing auxiliary OOD data during training, BiDO+ reduces BiDO's $\mu_{\mathrm{OOD}}$ from 0.60 to 0.22 on FaceScrub, while retaining most of its inversion resistance (PPA: 49.40\% vs.\ 52.39\%). This suggests that although OOD overconfidence contributes to BiDO's resistance, other properties of the defense continue to impede inversion. In contrast, Mixup achieves substantially stronger resistance (10.93\% PPA), without inducing elevated OOD confidence. Reconstruction distances agree with attack accuracy: on FaceScrub, Mixup's $\delta_{\mathrm{face}}$ is second only to Neg.LS (1.16 vs.\ 1.17), and on CelebA the methods with larger distances (MIDRE, LoFT) have 11--12 points lower test accuracy.

\subsection{Adversarial Robustness}
\label{sec:adversarial_robustness}

\citet{AT-AT} show that MIA defenses can gain inversion resistance at the expense of adversarial robustness. We therefore examine how much robustness Mixup sacrifices for the resistance it provides. Following~\citep{AT-AT}, we measure worst-case AutoAttack~\citep{autoattack} robust accuracy (RA) at $\ell_\infty$ budgets $\epsilon\in\{0.0025, 0.0005, 10^{-5}\}$, and define the cost of a method $M$ at budget $\epsilon$ as
\begin{equation}
\mathrm{Cost}_{\epsilon}(M)=\frac{100\,\bigl(\mathrm{RA}_{\epsilon}(\mathrm{Regular})-\mathrm{RA}_{\epsilon}(M)\bigr)}{\mathrm{PPA}(\mathrm{Regular})-\mathrm{PPA}(M)},
\label{eq:cost}
\end{equation}
\emph{i.e.}, the RA lost relative to the regular model per percentage point of PPA reduction; lower is better.

As shown in Tab.~\ref{tab:adversarial_robustness}, Mixup's cost is an order of magnitude lower than AT-AT's: 0.12-0.28 vs.\ 1.30-1.78 at $\epsilon=0.0005$ (8-12$\times$ less), and 0.01-0.06 vs.\ 0.62-0.98 at $\epsilon=10^{-5}$. This is not merely an effect of normalization: Mixup also loses far less RA in absolute terms (8.7-11.8 vs.\ 70.7-94.5 points at $\epsilon=0.0005$), while achieving a larger PPA reduction in every setting. Cost is omitted at $\epsilon=0.0025$, where even the regular model's RA is already low, and for AT-AT on CelebA/DenseNet-121, whose PPA reduction is below 10 points. Mixup thus provides stronger inversion resistance at a much lower robustness price.

\section{Conclusion}
\label{conclusion}



In this work, we revisit MIA defenses by examining not only how effectively they suppress inversion, but also why they work and what costs they impose. We find that BiDO, MIDRE, and Neg.LS share OOD overconfidence, which strongly tracks inversion resistance within each defense. Through confidence-landscape analysis and inversion trajectories, supported by temperature-scaling interventions, we show how elevated confidence on non-target inputs can cause premature attack saturation. However, this resistance comes at a reliability cost, as unseen individuals may be confidently assigned to known identities. Importantly, Input Mixup demonstrates that this trade-off is not inevitable, achieving the lowest attack accuracy in 7 of 8 settings without elevated OOD confidence or auxiliary OOD data, while limiting the maximum test-accuracy drop to 4.1\%.

\newpage

\section*{LLM Usage Statement}
We used an internally developed LLM tool to assist with language polishing. All authors are fully responsible for the content, results, and statements presented in this paper.

{\small
\bibliographystyle{ieeenat_fullname}
\bibliography{main}
}
\end{document}